\documentclass[conference]{IEEEtran}
\IEEEoverridecommandlockouts
\usepackage{cite}
\usepackage{amsmath,amssymb,amsfonts}
\usepackage{algorithmic}
\usepackage{graphicx}
\usepackage{textcomp}
\usepackage{xcolor}
\usepackage{subcaption}
\usepackage{graphics}
\usepackage{bm}
\usepackage{svg}
\usepackage{multirow}
\usepackage[ruled,vlined,linesnumbered]{algorithm2e}
\usepackage{url}
\usepackage{pifont}

\newcommand{\cmark}{\ding{51}}%
\newcommand{\xmark}{\ding{55}}

\newcommand*{\affaddr}[1]{#1}
\newcommand*{\affmark}[1][*]{\textsuperscript{#1}}
\newcommand*{\email}[1]{\texttt{#1}}

\DeclareMathOperator{\Conv1d}{Conv1d}
\DeclareMathOperator{\MaxPool}{MaxPool}
\DeclareMathOperator{\FFNN}{FFNN}
\DeclareMathOperator{\ReLU}{ReLU}
\DeclareMathOperator{\Softmax}{Softmax}
\DeclareMathOperator{\Tanh}{Tanh}
\DeclareMathOperator{\Concat}{Concat}
\DeclareMathOperator{\GRU}{GRU}
\DeclareMathOperator{\Sigmoid}{Sigmoid}
\DeclareMathOperator{\Norm}{Norm}

\def\BibTeX{{\rm B\kern-.05em{\sc i\kern-.025em b}\kern-.08em
    T\kern-.1667em\lower.7ex\hbox{E}\kern-.125emX}}
\begin{document}
\title{Cardiac Complication Risk Profiling for Cancer Survivors via Multi-View Multi-Task Learning}

\author{%
Thai-Hoang Pham\affmark[1,2], Changchang Yin\affmark[1,2], Laxmi Mehta\affmark[3], Xueru Zhang\affmark[1], and Ping Zhang\affmark[1,2]\\
\affaddr{\affmark[1]Department of Computer Science and Engineering, The Ohio State University, USA}\\
\affaddr{\affmark[2]Department of Biomedical Informatics, The Ohio State University, USA}\\
\affaddr{\affmark[3]Department of Medicine, Division of Cardiology, The Ohio State University, USA}\\
\email{\{pham.375,yin.371,mehta.149,zhang.12807,zhang.10631\}@osu.edu}\\
}

\maketitle

\begin{abstract}
Complication risk profiling is a key challenge in the healthcare domain due to the complex interaction between heterogeneous entities (e.g., visit, disease, medication) in clinical data. With the availability of real-world clinical data such as electronic health records and insurance claims, many deep learning methods are proposed for complication risk profiling. However, these existing methods face two open challenges. First, data heterogeneity relates to those methods leveraging clinical data from a single view only while the data can be considered from multiple views (e.g., sequence of clinical visits, set of clinical features). Second, generalized prediction relates to most of those methods focusing on single-task learning, whereas each complication onset is predicted independently, leading to suboptimal models. We propose a multi-view multi-task network (MuViTaNet) for predicting the onset of multiple complications to tackle these issues. In particular, MuViTaNet complements patient representation by using a multi-view encoder to effectively extract information by considering clinical data as both sequences of clinical visits and sets of clinical features. In addition, it leverages additional information from both related labeled and unlabeled datasets to generate more generalized representations by using a new multi-task learning scheme for making more accurate predictions. The experimental results show that MuViTaNet outperforms existing methods for profiling the development of cardiac complications in breast cancer survivors. Furthermore, thanks to its multi-view multi-task architecture, MuViTaNet also provides an effective mechanism for interpreting its predictions in multiple perspectives, thereby helping clinicians discover the underlying mechanism triggering the onset and for making better clinical treatments in real-world scenarios.

\end{abstract}


\section{Introduction}

Cardiovascular diseases are widely known as the leading causes of mortality in breast cancer survivors~\cite{schairer:2004,patnaik:2011,abdel:2019, strongman:2019}. With the recent substantial improvement of breast cancer survival rates, predicting the onset of multiple cardiac complications has become a critical task for enhancing patients' life quality. It is also a key to cost-effective disease management and prevention. However, this task is highly challenging because of the complex interactions between heterogeneous clinical entities. Effectively capturing these interactions may lead to more precise prediction and treatment for cancer survivors. 

\begin{figure}[t]
\centering
\includegraphics[width=0.85\linewidth]{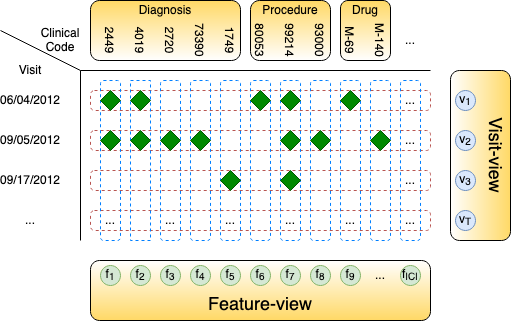}
\caption{Visit-view (sequence of clinical visits (rows)) and feature-view (set of clinical codes (columns)) of clinical data.} 
\label{fig:1}
\end{figure}

Over the past few decades, the rapid growth of real-world clinical data such as electronic health record (EHR) and insurance claims makes them valuable data sources used in data-driven (e.g., deep learning) systems for clinical risk prediction, especially complication risk profiling~\cite{ma:2017, baytas:2017, gao:2020}. As shown in Figure~\ref{fig:1}, this data includes heterogeneous clinical entities (e.g., visit, disease, medication) and can be considered from multiple views (i.e., sequence of visits, set of features). However, most existing studies consider each clinical outcome prediction separately and extract information in clinical data from a single view, thereby, making them not well-suited for complication risk profiling and raising two challenges.

\textbf{C1.} Clinical data is highly complex due to its heterogeneous and hierarchical structure. Thus, encoding patient records from single-view cannot provide comprehensive representations of these patients, and thereby cannot achieve superior prediction performance. In particular, by considering patient records as sequences of visits, previous works only learn the dependencies among clinical visits but cannot explicitly capture dynamic patterns of clinical features and their interaction at the global (i.e., sequence) level.

\textbf{C2.} Treating each complication onset prediction independently can lead to suboptimal models, especially in limited datasets. This is because it cannot capture the dependencies among complications that are manifestations caused by their common underlying condition. Moreover, this approach cannot exploit meaningful clinical patterns from unlabeled data, which is much easier to collect and can be used to improve prediction performance when labeled data is limited.

To tackle the two aforementioned challenges, we propose a new neural network-based framework named \underline{Mu}lti-\underline{Vi}ew Multi-\underline{Ta}sk \underline{Net}work (MuViTaNet) for cardiac complication risk profiling. This proposed model consists of a multi-view encoder (dealing with \textbf{C1}) and a novel multi-task learning (MTL) scheme (dealing with \textbf{C2}). In particular, the \textbf{multi-view encoder} includes visit-view and feature-view encoders that capture information from clinical visits and features simultaneously. The visit-view encoder considers a patient record as the sequence of clinical visits and captures the temporal relation among visits by Gated Recurrent Unit (GRU) network. The feature-view encoder considers the patient record as the set of temporal medical features, and then leverages convolutional neural networks (CNN) to extract temporal patterns from these features separately. Then, the max-pooling operation is applied to extract the most significant signals from temporal sequences. The \textbf{MTL scheme} utilizes an attention mechanism to learn complication-specific representation from shared information generated by the multi-view encoder. This scheme allows MuViTaNet to exploit additional information from related complications and unlabeled data to generate more generalized representations for the patient, which enables more accurate predictions. Moreover, by leveraging the attention mechanism associated multi-view encoder, the proposed model provides an efficient way to interpret its predictions from multiple perspectives, thereby helping clinicians discover the underlying mechanism triggering the onset and making better clinical treatments. 
We demonstrate that the proposed model significantly outperforms current state-of-the-art approaches for complication risk profiling task using multiple datasets derived from the insurance claim database. In summary, our contributions include the following:

\begin{itemize}
    \item We design a multi-view multi-task neural network architecture (MuViTaNet\footnote{Code is available at \url{https://github.com/pth1993/MuViTaNet}}) that accurately predicts multiple complication onsets and efficiently interprets its predictions.
    \item We develop a multi-view encoder to explicitly capture dependencies among clinical visits and clinical features from multiple views of clinical data.
    \item We also introduce a new MTL scheme that utilizes a complication-specific attention mechanism on top of the multi-view encoder to capture additional clinical information from related complications and unlabeled datasets.
    \item Finally, we conduct a comprehensive empirical study to demonstrate the effectiveness of MuViTaNet in terms of both prediction performance and interpretability compared to a wide range of previous approaches for cardiac complication risk profiling.
\end{itemize}

The remainder of the paper is organized as follows. Section II summarizes related works on clinical risk prediction in general and in particular, complication risk profiling. Section III describes the technical details of the proposed model (MuViTaNet). Section IV presents experimental results and discussions. Finally, Section V concludes the paper.

\section{Related Works}
In this section, we briefly review existing works related to our study including patient representation learning and MTL for clinical risk prediction, as well as complication risk profiling.

\textbf{Patient representation learning.} The abundance of real-world data in recent years creates an unprecedented opportunity to apply machine learning and data mining methods for clinical risk predictions. With the advancement of deep learning theory and the acceleration in computational technologies, neural network-based architectures can significantly improve prediction performance due to their ability to extract rich representations from data. Because of the temporal nature of clinical data, most existing methods rely on recurrent neural network architectures to learn patient representations, which are then used to make predictions for future clinical events (e.g., diagnosis, mortality, readmission, etc.)~\cite{ma:2017, baytas:2017, gao:2020, choi:2016, song:2018}. These works focused on designing attention mechanisms to capture dependencies among clinical visits~\cite{ma:2017, choi:2016, song:2018} and time-aware mechanisms to incorporate temporal information~\cite{baytas:2017, bai:2018, kwon:2018} into patient representation for making better predictions. Nonetheless, these models cannot explicitly capture the relationships among clinical features. Instead of considering EHR data as sequences of clinical visits, Concare~\cite{ma:2020} treats the record as the set of clinical features and extracts dynamic patterns of these features separately. Then the predictions are made by aggregating representations of all clinical features. However, all the existing methods only extract information from a single view of clinical data which makes the learned patient representations suboptimal. In contrast, we propose a multi-view model for capturing information from multiple views of clinical data simultaneously.

\textbf{Multi-task learning.} Multi-task learning (MTL) has been used widely across many applications of machine learning and data mining. By sharing information among related tasks, the prediction model can generalize better. In healthcare domain, some existing works applied MTL techniques to leverage information from related tasks to improve model performance in clinical risk prediction. In particular, both classical machine learning~\cite{zhou:2011, liu:2018, wiens:2016} and deep learning models~\cite{nori:2015, razavian:2016, lipton:2015} are formulated as MTL frameworks and are applied on a wide range of healthcare applications including disease progression modeling~\cite{zhou:2011}, mortality prediction~\cite{nori:2015}, disease onset prediction~\cite{razavian:2016}, and diagnosis classification~\cite{lipton:2015}.

\textbf{Complication risk profiling.} Mitigating the risk of complications is crucial for many disease management programs. Despite its importance, there have not been many existing methods designed for this task. Unlike a single clinical risk prediction task, complication risk profiling requires multiple predictions for onset of complications. Thus, capturing relationships among related complications is crucial to achieving good prediction performances. Some methods have been proposed to predict the onset of complications of some diseases and clinical procedures. For example, multi-task logistic regression has been used to predict complication risks for diabetes care~\cite{liu:2018, liu:2019}. Besides linear models, the deep learning method is also used to predict complications of this chronic disease~\cite{ljubic:2020} but this work considers each complication independently. For breast cancer survivors, relationships between cardiac complications and cancer were also investigated~\cite{guo:2020, strongman:2019, abdel:2019} to show the correlation between these two diseases.

\section{Methodology}

\begin{figure}[t]
\centering
\subcaptionbox{}%
  [.17\linewidth]{\includegraphics[width=\linewidth]{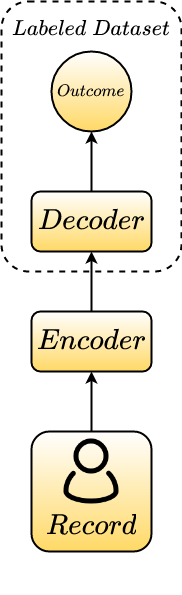}}\hfill
\centering
\subcaptionbox{}
  [.3\linewidth]{\includegraphics[width=\linewidth]{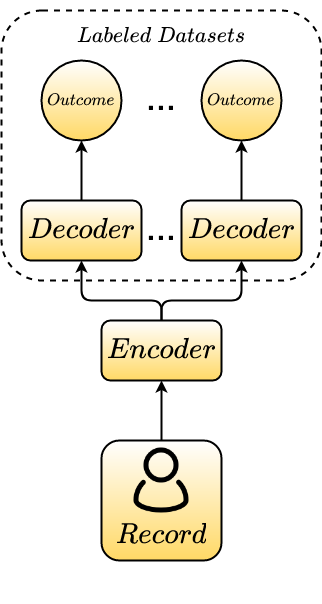}}\hfill
\centering
\subcaptionbox{}
  [.5\linewidth]{\includegraphics[width=\linewidth]{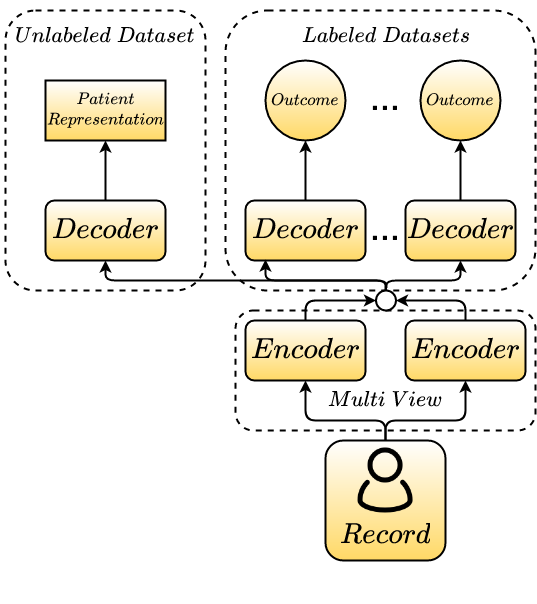}}
 \caption{General schemes for learning from clinical data. (a) single-view single-task learning, (b) single-view multi-task learning, (c) multi-view multi-task learning. Our proposed model belongs to multi-view multi-task learning with the multi-view encoder (i.e., visit-view and feature-view) and the task-specific attention mechanisms and decoders for both labeled and unlabeled datasets.}
\label{fig:2}
\end{figure}

\begin{figure*}[t]
\centering
\includegraphics[width=0.9\linewidth]{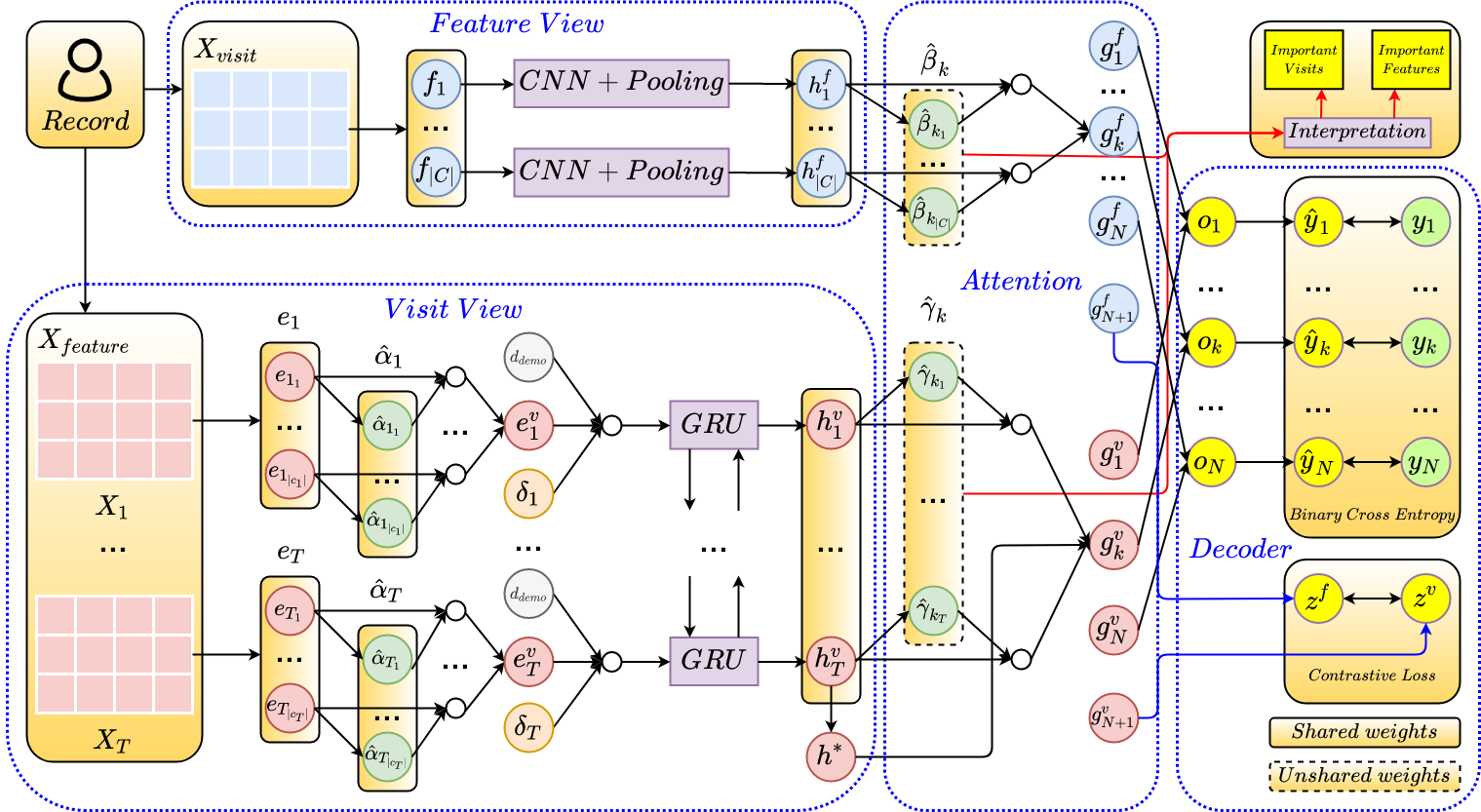}
 \caption{The overall architecture of MuViTaNet. The proposed framework consists of four main components: feature-view encoder, visit-view encoder, task-specific attention, and task-specific decoder. Given a patient record, MuViTaNet first extracts information from clinical visits and features by looking at the record in two different ways: sequence of clinical visits and set of clinical features. Then the shared representation learned by these two encoders is put into the task-specific attention to learn the task-specific representation. Finally, the clinical predictions are generated by the task-specific decoders. Note that the figure only shows the task-specific attention for one prediction task for simplicity.}  
\label{fig:3}
\end{figure*}

In this section, we first give brief introduction about patient records, complication risk profiling task and the corresponding notations. Then, we present our proposed model MuViTaNet.

\subsection{Definitions and Basic Notations}

Definitions and notations used in this study are shown in the following paragraphs and are summarized in Table~\ref{tab:1}.

\textbf{Patient Records.}
The heterogeneous and hierarchical structure of a patient record is defined as follows.
\begin{itemize}
    \item \textbf{Definition 1 (Clinical Code).} $\bm{C} = \{c_{1}, c_{2},\cdots,c_{|\bm{C}|}\}$ is the set of unique clinical codes including diagnosis, procedure, and medication codes with $|\bm{C}|$ is the number of these unique codes. Each code $c_{i}$ can be represented by binary vector $\bm{x}_{i} \in \{0,1\}^{|\bm{C}|}$ where $i^{th}$ element of this vector is 1 and other elements are 0.
    \item \textbf{Definition 2 (Clinical Visit).} An visit is a hospital stay from admission to discharge. Each visit $\bm{v}_{j}$ is a tuple of $(\bm{c}_{j}$, $t_{j})$ where $\bm{c}_{j} = \{ c_{j_{1}}, c_{j_{2}}, \cdots, c_{j_{|\bm{c}_{j}|}} \} \in \bm{C}^{|\bm{c}_{j}|}$ with set of indexes $\{j_{1}, \cdots, j_{|\bm{c}_{j}|}\} \in \{1, 2, \cdots, |\bm{C}|\}$ and $t_{j}$ is the timestamp of the visit. $\bm{c}_{j}$ can be represented by binary vector $\bm{V}_{j} \in \{0,1\}^{|\bm{C}|}$ where the $i^{th}$ element is 1 if $\bm{c}_{j}$ contains the code $c_{i}$. Besides vector representation, $\bm{c}_{j}$ can also be expressed as matrix $\bm{X}_{j} \in \{0,1\}^{|\bm{c}_{j}| \times |\bm{C}|}$ 
    where $i^{th}$ row of this matrix is the binary vector $\bm{x}_{j_{i}} \in \{0,1\}^{|\bm{C}|}$ of code $c_{j_{i}}$. 
    \item \textbf{Definition 3: (Patient Record).} The patient record $\bm{P}$ is a sequence of visits $[\bm{v}_{1}, \bm{v}_{2}, \cdots, \bm{v}_{T}]$ where $T$ is the number of visits. Like clinical visit representation, $\bm{P}$ can be represented at the two different granularities. At visit-level, $\bm{P}$ can be represented as a binary matrix $\bm{X}_{visit} \in \{0,1\}^{T \times |\bm{C}|}$ where $j^{th}$ row of this matrix is binary vector $\bm{V}_{j}$ of visit $\bm{v}_{j}$. At feature-level, $\bm{P}$ can be represented as the sequence of matrices $\bm{X}_{feature} = [\bm{X}_{1}, \bm{X}_{2}, \cdots, \bm{X}_{T}]$.
    \item \textbf{Definition 4: (Demographic Information).} Besides clinical information, a patient record can have demographic information about the patient such as age, gender, region, etc. It can be represented by binary vector $\bm{d}_{demo} \in \{0,1\}^{d_{demo}}$, where $d_{demo}$ is the number of demographic attributes.
\end{itemize}

\textbf{Clinical Risk Profiling.}
The aim of this task is to find a set of functions $\bm{F} = \{ F_{1}, F_{2}, \cdots, F_{N} \}$ that predicts the onset of complications $\bm{Y} \in \mathbb{R}^{N}$ from patient record $\bm{P}$, where $N$ is the number of complications. In MTL setting, $F_{1}, F_{2}, \cdots, F_{N}$ generally have some shared parameters to learn shared information from related tasks for better predictions.

\begin{table}[t]
\caption{Notation definition}
\resizebox{0.48\textwidth}{!}{
\begin{tabular}{ll}
\hline
Notation                                                                               & Description                                       \\ \hline
$\bm{C}$                                                                                      & Set of clinical codes/features                              \\ 
$\bm{P}$                                                                                      & A patient record                                  \\ 
$c_{i}$                                                                               & $i^{th}$ clinical codes in set C                       \\ 
$\bm{x}_{i} \in \{0,1\}^{|\bm{C}|}$                                 & vector representation of code $c_{i}$     \\ 
$\bm{v}_{j}$                                                                               & $j^{th}$ clinical visit in $\bm{P}$                                   \\ 
$\bm{c}_{j}$                                                                              & set of clinical codes in visit $\bm{v}_{j}$            \\ 
$t_{j}$                                                                               & timestamp of visit $\bm{v}_{j}$                       \\ 
$\bm{V}_{j} \in \{0,1\}^{|\bm{C}|}$                                 & vector representation of visit $\bm{v}_{i}$    \\ 
$\bm{X}_{j} \in \{0,1\}^{|\bm{c}_{i}| \times |\bm{C}|}$ & matrix representation of $\bm{v}_{j}$    \\ 
$\bm{X}_{visit} \in \{0,1\}^{T \times |\bm{C}|}$      & visit-level representation of $\bm{P}$                   \\ 
$\bm{X}_{feature} \in T \times (\{0,1\}^{|\bm{c}_{i}| \times |\bm{C}|})$                                                                            & feature-level representation of $\bm{P}$                    \\ 
$\bm{d}_{demo}$                                                                             & vector representation of demographics      \\ 
$\bm{\widehat{\alpha}}_{j} \in \mathbb{R}^{|\bm{c}_{j}|}$                                                                               & attention weights of codes in visit $\bm{v}_{j}$   \\ 
$\bm{\widehat{\beta}}_{j} \in \mathbb{R}^{|\bm{c}_{i}|}$                                                                               & task-specific attention weights for features  \\ 
$\bm{\widehat{\gamma}}_{j} \in \mathbb{R}^{T}$                                                           & task-specific attention weights for visits \\ 
$\bm{\delta}_{i} \in \mathbb{R}^{d}$                                                           & temporal encoding vector of visit $\bm{v}_{i}$        \\ 
$\bm{H}^{v} \in \mathbb{R}^{T \times 2d}$                                                               & representation learned by visit-view encoder               \\ 
$\bm{h}^{*} \in \mathbb{R}^{2d}$                                                               & patient representation             \\
$\bm{H}^{f} \in \mathbb{R}^{|\bm{C}| \times 4d}$                                                               & representation learned by feature-view encoder             \\
$\bm{g}^{v}_{k} \in \mathbb{R}^{2d}$ & visit-view task-specific representation for $k^{th}$ task \\
$\bm{g}^{f}_{k} \in \mathbb{R}^{4d}$ & feature-view task-specific representation for $k^{th}$ task \\
$\bm{o}_{k} \in \mathbb{R}^{8d}$                                                                               & task-specific representation for $k^{th}$ task        \\ 
$y_{k}$ & ground-truth output for $k^{th}$ task \\
$\widehat{y}_{k}$ & predicted output for $k^{th}$ task \\
\hline
\end{tabular}}
\label{tab:1}
\end{table}

\begin{algorithm}[t]
\SetAlgoLined
\KwIn{Datasets $\{ \bm{D}_{k} \}_{k=1}^{N+1}$, set of clinical codes $\bm{C}$, batch sizes $n_{s}$, $n_{u}$}
\KwOut{Trained model parameters $\bm{\theta} = \{\bm{\theta}^{shared}, \{ \bm{\theta}^{task-specific}_{k} \}_{k=1}^{N}\}$}
Randomly initialize $\bm{\theta}$\;
Calculate sampling rate for each dataset $\lambda_{k} = \frac{|\bm{D}_{k}| / n_{k}}{\sum_{{k}'=1}^{N} |\bm{D}_{{k}'}| / n_{{k}'}}$ ($n_{k} = n_{u}$ if $k = N+1$, $n_{k} = n_{s}$ otherwise)\;
\For{epoch $=1$ to E}{
    \Repeat{$\{ \bm{D}_{k} \}_{k=1}^{N+1} = \varnothing$}{
    Select dataset $\bm{D}_{k} \sim \bm{\lambda}$\;
    Initialize loss $L_{k} = 0$\;
    Select sample batch $\bm{b}$ from dataset $\bm{D}_{i}$\;
    \For{patient $\bm{P}_{i}$ in batch $\bm{b}$}{
        $(\bm{X}_{feature}, \bm{X}_{visit}) = \bm{P}_{i}$\;
        Obtain feature-view representation $\bm{H}^{f}$ from $\bm{X}_{visit}$ using Eq.~(\ref{eq:1}), (\ref{eq:2})\;
        Obtain visit-view representation $\bm{H}^{v}$ and patient representation $\bm{h}^{*}$ from $\bm{X}_{feature}$ using Eq.~(\ref{eq:3})-(\ref{eq:11})\;
        Calculate task-specific attention weights $\bm{\widehat{\beta}}, \bm{\widehat{\gamma}}$ from $\bm{H}^{f}$, $\bm{H}^{v}$ using Eq.~(\ref{eq:12})\;
        Obtain task-specific representations using Eq.~(\ref{eq:13})\;
        \uIf{$k \in \{1,\cdots,N\}$}{
            Calculate prediction $\hat{y}_{k_{i}}$ using Eq.~(\ref{eq:14})\;
            Calculate BCE loss $L_{k_{i}}$ using Eq.~(\ref{eq:16})\;}
        \uElse{
            Project multi-view representations to unit hypersphere using Eq.~(\ref{eq:15})\;
            Calculate CL loss $L_{k_{i}}$ using Eq.~(\ref{eq:17})\;
        }
        $L_{k} = L_{k} + L_{k_{i}}$\;
    }
    Update parameters $\bm{\theta}$ using gradient of $L_{k}$\;
    $\bm{D}_{k} = \bm{D}_{k} \setminus \bm{b}$\;
    }
}
\caption{Training procedure for MuViTaNet}
\label{alg:1}
\end{algorithm}

\subsection{Proposed Model}
\textbf{Overview Architecture.}
This section presents our proposed multi-view multi-task network (MuViTaNet) for predicting onset of multiple complications from patient records. MuViTaNet is designed to explicitly capture the dependencies among clinical visits and clinical features from patient records. It leverages additional information from both related labeled and unlabeled data in MTL to achieve accurate predictions and efficient interpretation. In particular, MuViTaNet consists of four main components as follows. (1) \underline{Feature-view Encoder}. This component considers a patient record as a set of temporal clinical features and then encodes information of each feature separately. (2) \underline{Visit-view Encoder}. This component formulates a patient record as a sequence of visits and then learns a representation for each visit in the sequential context. Specifically, this component is designed as a hierarchical model that exploits patient records in the two levels, including feature-level and visit-level. (3) \underline{Task-specific Attention}. After learning the shared representation from feature-view and visit-view encoders, an attention mechanism is employed to extract task-specific representation for each task from the shared representation. (4) \underline{Task-specific Decoder}. The task-specific representations are fed into the corresponding task-specific decoders to predict clinical outcomes for patients in complication datasets and to project representations to unit hypersphere for patients in unlabeled dataset. Figure~\ref{fig:3} shows the overview architecture of MuViTaNet and technical details of its components are presented as follows.


\textbf{Feature-view Encoder.}
This component treats patient data as a set $\bm{C}$ of clinical codes which are represented by the set of temporal sequences (i.e., columns of matrix $\bm{X}_{visit} \in \{0,1\}^{T \times |\bm{C}|}$). In particular, given clinical code $c_{i}$, its temporal data can be represented by a binary vector $\bm{f}_{i} \in \{0,1\}^{T}$ which is $i^{th}$ column of $\bm{X}_{visit}$. Then, one-dimensional convolutional neural networks (Conv1d) and max-pooling (MaxPool) operation are employed to extract temporal patterns from each clinical code separately. In particular, Conv1d with kernel size $k$ (i.e., $k=3$ in our setting) takes as inputs the sub-sequences of length k from vector $\bm{f}_{i}$ to learn the representation of code $c_{i}$ as follows.
\begin{equation}
    \bm{H}_{i}^{f} = \Conv1d(\bm{f}_{i})
\label{eq:1}
\end{equation}
where $\bm{H}_{i}^{f} \in \mathbb{R}^{4d \times T}$ are the output of $\Conv1d$ and $4d$ is the number of filters used in convolution operations.
Next, the row-wise max-pooling is applied to $\bm{H}_{i}^{f}$ to generate vector representation for clinical code $c_{i}$.
\begin{equation}
    \bm{h}_{i}^{f} = \MaxPool(\bm{H}_{i}^{f})
\label{eq:2}
\end{equation}
Note that the weights of Conv1d are not shared between clinical codes. The output of feature-view encoder is matrix $\bm{H}^{f} = [\bm{h}_{1}^{f}, \bm{h}_{2}^{f}, \cdots, \bm{h}_{|\bm{C}|}^{f}] \in \mathbb{R}^{|\bm{C}| \times 4d}$.

\textbf{Visit-view Encoder.}
This component formulates patient data as a sequence of visits in which each visit can be seen as a set of clinical codes. Due to the hierarchical characteristic of this data structure, the visit-view encoder is also designed hierarchically to capture information at different levels. Given visit $\bm{v}_{j}$, we represent this visit by matrix $\bm{X}_{j} \in \{0,1\}^{|\bm{c}_{j}| \times |\bm{C}|}$ which is $j^{th}$ element of the sequence $\bm{X}_{feature}$. Because different clinical codes associated with the same visit can have disparate impacts, instead of treating these clinical codes uniformly when aggregating them to represent the visit, the location attention mechanism is employed to learn the contributions of these clinical codes to their visit representation. In particular, given a binary representation $\bm{x}_{j_{i}} \in \bm{X}_{j}$ of code $c_{j_{i}}$, 1-layer feed-forward neural network is applied to learn the dense representation from sparse vector of this clinical code as follows.

\begin{equation}
    \bm{e}_{j_{i}} = \FFNN_{1}(\bm{x}_{j_{i}}) = \ReLU(\bm{W}_{1}\bm{x}_{j_{i}} + \bm{b}_{1})
\label{eq:3}
\end{equation}
where $\bm{W}_{1} \in \mathbb{R}^{d \times |\bm{C}|}$ is the learned weight matrix of clinical codes, $\bm{b}_{1} \in \mathbb{R}^{d}$ is the bias vector, and ReLU is rectified linear unit activation function. Then the 2-layer feed-forward neural network $\FFNN_{2}$ with $\Tanh$ activation function is used to generate the attention score $\alpha_{j_{i}}$ for this clinical code as follows.
\begin{equation}
    \alpha_{j_{i}} = \FFNN_{2}(\bm{e}_{j_{i}})
\label{eq:4}
\end{equation}
The attention vector $\bm{\alpha}_{j} = [\alpha_{j_{1}}, \alpha_{j_{2}}, \cdots, \alpha_{j_{|\bm{c}_{j}|}}]$ which represents the contributions of clinical codes in visit $\bm{v}_{j}$ is fed into the softmax layer to get the normalized vector $\bm{\widehat{\alpha}}_{j} = [\widehat{\alpha}_{j_{1}}, \widehat{\alpha}_{j_{2}}, \cdots, \widehat{\alpha}_{j_{|\bm{c}_{j}|}}] \in \mathbb{R}^{|\bm{c}_{j}|}$.
\begin{equation}
    \bm{\widehat{\alpha}}_{j} = \Softmax(\bm{\alpha}_{j})
\label{eq:5}
\end{equation}
Then, the representation of visit $\bm{v}_{j}$ are computed as the weighted average of its clinical codes.
\begin{equation}
    \bm{e}_{j}^{v} = (\bm{\widehat{\alpha}}_{j})^{T} \bm{e}_{j}
\label{eq:6}
\end{equation}
where $\bm{e}_{j} = [\bm{e}_{j_{1}}, \bm{e}_{j_{2}}, \cdots, \bm{e}_{j_{|\bm{c}_{j}|}}] \in \mathbb{R}^{|\bm{c}_{j}| \times d}$ denotes the $j^{th}$ visit's representation.
To generate personalized representation for each visit, demographic information including age and region is incorporated into every clinical visit as follows.
\begin{equation}
    \ddot{\bm{e}}_{j}^{v} = \bm{W}_{2}(\Concat(\bm{e}_{j}^{v}, \bm{d}_{demo}))
\label{eq:7}
\end{equation}
where $\Concat$ is the concatenation operation and $\bm{W}_{2} \in \mathbb{R}^{(d + d_{demo}) \times d}$ is the weight matrix mapping concatenated vectors to the original embedding space. Besides clinical codes, each visit is also associated with its timestamp. In order to capture the elapsed time between visits, we add the temporal encoding vector to each visit as follows.
\begin{align}
    \bm{\widehat{e}}_{j}^{v} = \ddot{\bm{e}}_{j}^{v} + \bm{\delta}_{j}
\label{eq:8}
\end{align}
where $\bm{\delta}_{j} \in \mathbb{R}^{d}$ is the temporal encoding vector whose design is inspired by the positional encoding used in Transformer architecture~\cite{vaswani:2017}. In particular, it is computed by trigonometric functions as follows.
\begin{equation}
\begin{aligned}
   \delta_{j, 2t} &= \sin \left (\frac{t_{T} - t_{j}}{10000^{2t/d}} \right ) \\ 
   \delta_{j, 2t+1} &= \cos \left (\frac{t_{T} - t_{j}}{10000^{2t/d}} \right )  
\end{aligned}
\label{eq:9}
\end{equation}
where $0 \leq 2t < d-1$. From Equation~\eqref{eq:9}, we can see that temporal embedding encodes similar time intervals into similar vectors in embedding space. 

To generate the sequential representations for visits in the sequential context, we put the independent representations for visits learned from previous steps into the bidirectional GRU layer. Specifically, the sequential representation for these visits is computed as follows.
\begin{equation}
\begin{aligned}
    \overrightarrow{\textbf{h}_{j}} &= \GRU(\widehat{\textbf{e}}_{j}^{v}, \overrightarrow{\textbf{h}_{j-1}}) \\
    \overleftarrow{\textbf{h}_{j}} &= \GRU(\widehat{\textbf{e}}_{j}^{v}, \overleftarrow{\textbf{h}_{j+1}}) \\
    \textbf{h}_{j}^{v} &= \Concat(\overrightarrow{\textbf{h}_{j}}, \overleftarrow{\textbf{h}_{j}})
\end{aligned}
\label{eq:10}
\end{equation}
where $ \bm{h}_{j}^v \in \mathbb{R}^{2d}$. Then, the patient representation is computed based on the last visit in the visit sequence.
\begin{equation}
    \bm{h}^{*} = \FFNN_{3}(\bm{h}_{T}^{v})
\label{eq:11}
\end{equation}
In summary, the outputs of the visit-view encoder include the sequential representations of clinical visits $\bm{H}^{v} = [\bm{h}_{1}^{v}, \bm{h}_{2}^{v}, \cdots, \bm{h}_{T}^{v}] \in \mathbb{R}^{T \times 2d}$ and the patient representation $\bm{h}^{*} \in \mathbb{R}^{2d}$.

\textbf{Task-specific Attention.}
Given the shared representations generated by feature-view and visit-view encoders, attention mechanisms are employed to generate the task-specific representations for the patient. Specifically, the attention weights of clinical features and visits for $k^{th}$ task are computed as follows.

\begin{equation}
\begin{aligned}
\beta_{k_{i}} &= \FFNN^{k}_{4}(\bm{h}_{i}^{f}) \\
\gamma_{k_{j}} &= \FFNN^{k}_{5}(\bm{h}_{j}^{v}) \\
\bm{\widehat{\beta}}_{k} &= \Softmax([\beta_{k_{1}}, \beta_{k_{2}}, \cdots, \beta_{k_{|\bm{C}|}|}]) \\
\bm{\widehat{\gamma}}_{k} &= \Softmax([\gamma_{k_{1}}, \gamma_{k_{2}}, \cdots, \gamma_{k_{T}}])
\end{aligned}
\label{eq:12}
\end{equation}
where $\FFNN^{k}_{4}, \FFNN^{k}_{5}$ are 2-layer feed-forward neural networks with $\Tanh$ activation function that compute the weights of clinical features and visits from their representations. Then, we obtain the task-specific representation $\bm{o}_{k} \in \mathbb{R}^{8d}$ for $k^{th}$ task as follows.
\begin{equation}
\begin{aligned}
    \bm{g}^{f}_{k} &= (\bm{\widehat{\beta}}_{k})^{T} \bm{H}^{f} \\
    \bm{g}^{v}_{k} &= (\bm{\widehat{\gamma}}_{k})^{T} \bm{H}^{v} \\
    \bm{o}_{k} &= \Concat(\bm{g}^{f}_{k}, \bm{g}^{v}_{k}, \bm{h}^{*})
\end{aligned}
\label{eq:13}
\end{equation}
\textbf{Task-specific Decoder.}
For a patient in labeled dataset (i.e., complication dataset), the 2-layer feed forward neural network with $\Sigmoid$ activation function at the last layer is employed to predict the probability of complication onset for this patient.
\begin{equation}
    \hat{y}_{k} = \FFNN^{k}_{6}(\bm{o}_{k}), \;\;\;\; k \in \{1,\cdots,N\}
\label{eq:14}
\end{equation}
For a patient in unlabeled dataset, the 2-layer feed forward neural network with normalization operation ($\Norm$) is used to project the feature-view and visit-view representations of this patient on the unit hypersphere.
\begin{equation}
\begin{aligned}
    \bm{z}^{f} &= \Norm(\FFNN^{k}_{6}(\bm{g}^{f}_{k})), \;\;\;\; k = N+1 \\
    \bm{z}^{v} &= \Norm(\FFNN^{k}_{6}(\Concat(\bm{g}^{v}_{k}, \bm{h}^{*}))) \\
\end{aligned}
\label{eq:15}
\end{equation}

\textbf{Optimization.}
To train MuViTaNet in MTL setting, we follow the alternating training strategy~\cite{dong:2015} in which each task is selected randomly and then is optimized for a fixed number of parameter updates before switching to other tasks. In our setting, different tasks have datasets of different sizes,  so we select a task to optimize with probability $\lambda_{k} = \frac{|\bm{D}_{k}| \setminus n_{k}}{\sum_{{k}'=1}^{N+1} |\bm{D}_{{k}'}| \setminus  n_{{k}'}}$, where $\bm{D}_{k}$ and $n_{k}$ are the dataset and batch size for $k^{th}$ task, and $N$ is the number of complication datasets.

For labeled datasets, the binary cross-entropy (BCE) loss function is used to optimize the prediction based on ground-truth labels. Specifically, for $k^{th}$ task with dataset $\bm{D}_{k}$, the loss function for this task is computed as follows.
\begin{equation}
    L^{k}_{L} = -\frac{1}{|D_{k}|} \sum_{i=1}^{|D_{k}|} \Big(y_{k_{i}}\log(\widehat{y}_{k_{i}}) + (1 - y_{k_{i}})\log(1 - \widehat{y}_{k_{i}}) \Big)
\label{eq:16}
\end{equation}
where $\bm{y}_{k}$ and $\bm{\widehat{y}}_{k}$ are the ground-truth and predicted outputs for $k^{th}$ task respectively. For unlabeled dataset, we leverage the contrastive (CL) loss function~\cite{chen:2020} to pull together the normalized representations of feature-view and visit-view of the same patient and to push apart these representations from representations of other patients.
\begin{equation}
    L_{U} = - \sum_{i=1}^{|D_{k}|} \sum_{\bm{z}_{i} \in \{ \bm{z}_{i}^{f}, {\bm{z}_{i}^{v} \}}}  \log \frac{\exp(\bm{z}_{i}^{f} \cdot \bm{z}_{i}^{v})}{\sum_{\bm{z}_{j} \in A(\bm{z}_{i})} \exp(\bm{z}_{i} \cdot \bm{z}_{j})}
\label{eq:17}
\end{equation}
where $A(\bm{z}_{i}) \equiv \bm{Z} \setminus \bm{z}_{i}$ in which $\bm{Z} = \{\bm{z}_{i}^{f}, \bm{z}_{i}^{v}\}_{i=1}^{|D_{k}|}$.


\section{Experiments}

\begin{table}[t]
\caption{Cardiac complications in female breast cancer cohort and their corresponding ICD codes and numbers of positive instances.}
\resizebox{0.5\textwidth}{!}{
\begin{tabular}{llll}
\hline
complication                 & Description                                                                                                                          & ICD-10 Codes                                                & \#subjects \\ \hline
Atrial Fibrillation         & \begin{tabular}[c]{@{}l@{}}An irregular, often rapid heart \\ rate that commonly causes \\ poor blood flow\end{tabular}              & I48                                                         & 322        \\ \hline
\begin{tabular}[c]{@{}l@{}}Coronary Artery \\ Disease\end{tabular}     & \begin{tabular}[c]{@{}l@{}}Damage or disease in the \\ heart's major blood vessels\end{tabular}                                      & I20-I25                                                     & 769        \\ \hline
Heart failure              & \begin{tabular}[c]{@{}l@{}}A chronic condition in which \\ the heart doesn't pump blood \\ as well as it should\end{tabular}         & \begin{tabular}[c]{@{}l@{}}I11, I13\\ I42, I50\end{tabular} & 1124       \\ \hline
Hypertension                & \begin{tabular}[c]{@{}l@{}}A condition in which the force \\ of the blood against the artery \\ walls is too high\end{tabular}       & I10, I16                                                    & 6787       \\ \hline
\begin{tabular}[c]{@{}l@{}}Peripheral Arterial \\ Disease\end{tabular} & \begin{tabular}[c]{@{}l@{}}A circulatory condition in \\ which narrowed blood vessels \\ reduce blood flow to the limbs\end{tabular} & I70                                                         & 340        \\ \hline
Stroke                      & \begin{tabular}[c]{@{}l@{}}Damage to the brain from \\ interruption of its blood supply\end{tabular}                                 & I60-I69                                                     & 592        \\ \hline
\end{tabular}}
\label{tab:2}
\end{table}

\renewcommand{\arraystretch}{1.5}
\begin{table*}[t]
\caption{Comparison of prediction performance measured by AU-ROC scores on six complication risk profiling tasks. We report the average AU-ROC scores and their corresponding standard deviation. (AF: Atrial Fibrillation, CAD: Coronary Artery Disease, HF: Heart Failure, PAD: Peripheral Arterial Disease).}
\resizebox{\textwidth}{!}{
\begin{tabular}{c|ll|ccccccc}
\cline{1-10}
\multicolumn{3}{c}{Method}                & AF                                        & CAD                                       & HF                                        & Hypertension                              & PAD                                       & Stroke                                    & Average                                   \\ \cline{1-10} 
\multirow{8}{*}{\rotatebox[origin=c]{90}{Single-task}}                 & \multirow{2}{*}{Classical}       & LR             & $0.6133 \pm 0.0437$                           & $0.6402 \pm 0.0165$                           & $0.6982 \pm 0.0088$                           & $0.7901 \pm 0.0088$                           & $0.5700 \pm 0.0341$                           & $0.6150 \pm 0.0128$                           & $0.6545 \pm 0.0208$                           \\
                            &                                        & RF             & $0.7159 \pm 0.0434$ & $0.7187 \pm 0.0260$                           & $0.7863 \pm 0.0147$                           & $0.8066 \pm 0.0090$                           & $0.6880 \pm 0.0525$                           & $0.7172 \pm 0.0262$                           & $0.7388 \pm 0.0286$                           \\ \cline{2-10}
                            & \multirow{2}{*}{\begin{tabular}[c]{@{}l@{}} Recurrent-\\based \end{tabular}} & GRU            & $0.6701 \pm 0.0425$                           & $0.7218 \pm 0.0116$                           & $0.7805 \pm 0.0033$                           & $0.8122 \pm 0.0084$                           & $0.6884 \pm 0.0368$                           & $0.7213 \pm 0.0103$                           & $0.7324 \pm 0.0188$                           \\
                            &                                        & Bi-GRU         & $0.6620 \pm 0.0533$                           & $0.7295 \pm 0.0079$                           & $0.7845 \pm 0.0058$                           & $0.8155 \pm 0.0098$                           & $0.6967 \pm 0.0172$ & $0.7291 \pm 0.0088$ & $0.7362 \pm 0.0171$                           \\ \cline{2-10}
                            & Time-aware                       & T-LSTM         & $0.6739 \pm 0.0518$                           & $0.7052 \pm 0.0133$                           & $0.7651 \pm 0.0156$                           & $0.8024 \pm 0.0118$                           & $0.6802 \pm 0.0239$                           & $0.6994 \pm 0.0203$                           & $0.7210 \pm 0.0228$                           \\ \cline{2-10}
        & \multirow{4}{*}{\begin{tabular}[c]{@{}l@{}} Attention-\\based \end{tabular}} & Dipole         & $0.6804 \pm 0.0661$                           & $0.7287 \pm 0.0120$                           & $0.7791 \pm 0.0026$                           & $0.8157 \pm 0.0081$                           & $0.6839 \pm 0.0320$                           & $0.7247 \pm 0.0040$                           & $0.7354 \pm 0.0209$                           \\
                            &                                        & RETAIN         & $0.6493 \pm 0.0465$                           & $0.6780 \pm 0.0196$                           & $0.7360 \pm 0.0139$                           & $0.8078 \pm 0.0086$                           & $0.6731 \pm 0.0224$                           & $0.6770 \pm 0.0112$                           & $0.7035 \pm 0.0126$                           \\
                            &                                        & Transformer    & $0.6516 \pm 0.0563$                           & $0.7021 \pm 0.0155$                           & $0.7502 \pm 0.0069$                           & $0.8107 \pm 0.0076$                           & $0.6721 \pm 0.0392$                           & $0.6981 \pm 0.0135$                           & $0.7141 \pm 0.0183$                           \\
                            &                                        & LSAN           & $0.6069 \pm 0.0556$                           & $0.6910 \pm 0.0135$                           & $0.7567 \pm 0.0180$                           & $0.8163 \pm 0.0085$                           & $0.6464 \pm 0.0464$                           & $0.6897 \pm 0.0206$                           & $0.7012 \pm 0.0271$                           \\ \cline{1-10}
\multirow{8}{*}{\rotatebox[origin=c]{90}{Multi-task}} & \multirow{2}{*}{\begin{tabular}[c]{@{}l@{}} Recurrent-\\based \end{tabular}}                                       & GRU         & $0.7915 \pm 0.0475$                           & $0.7759 \pm 0.0144$                           & $0.8186 \pm 0.0136$                           & $0.8143 \pm 0.0096$                           & $0.7524 \pm 0.0253$                           & $0.7458 \pm 0.0222$                           & $0.7831 \pm 0.0221$                           \\
                            &                                        & Bi-GRU      & $0.7984 \pm 0.0524$                           & $0.7824 \pm 0.0121$                           & $0.8279 \pm 0.0125$                           & $0.8189 \pm 0.0100$                           & $0.7503 \pm 0.0189$                           & $0.7462 \pm 0.0237$                           & $0.7873 \pm 0.0216$                           \\ \cline{2-10}
                            & Time-aware                                       & T-LSTM      & $0.7944 \pm 0.0466$                           & $0.7591 \pm 0.0093$                           & $0.8134 \pm 0.0124$                           & $0.8106 \pm 0.0087$                           & $0.7382 \pm 0.0285$                           & $0.7419 \pm 0.0232$                           & $0.7763 \pm 0.0214$                           \\ \cline{2-10}
                            & \multirow{4}{*}{\begin{tabular}[c]{@{}l@{}} Attention-\\based \end{tabular}}                                       & Dipole      & $0.7823 \pm 0.0620$                           & $0.7814 \pm 0.0213$                           & $0.8239 \pm 0.0095$                           & $0.8210 \pm 0.0092$                           & $0.7554 \pm 0.0350$                           & $0.7611 \pm 0.0194$                           & $0.7875 \pm 0.0261$                           \\
                            &                                        & RETAIN      & $0.7686 \pm 0.0485$                           & $0.7554 \pm 0.0083$                           & $0.8024 \pm 0.0165$                           & $0.8029 \pm 0.0066$                           & $0.7312 \pm 0.0263$                           & $0.7376 \pm 0.0254$                           & $0.7661 \pm 0.0219$                           \\
                            &                                        & Transformer & $0.7697 \pm 0.0649$                           & $0.7738 \pm 0.0110$                           & $0.8049 \pm 0.0164$                           & $0.8092 \pm 0.0106$                           & $0.7484 \pm 0.0423$                           & $0.7643 \pm 0.0083$                           & $0.7784 \pm 0.0256$                           \\
                            &                                        & LSAN        & $0.7775 \pm 0.0576$                           & $0.7788 \pm 0.0225$                           & $0.8082 \pm 0.0150$                           & $0.8226 \pm 0.0061$                           & $0.7599 \pm 0.0319$                           & $0.7533 \pm 0.0147$                           & $0.7834 \pm 0.0246$                           \\ \cline{2-10}
                            & Ours                                       & MuViTaNet      & $\bm{0.8120 \pm 0.0457}$ & $\bm{0.8070 \pm 0.0147}$ & $\bm{0.8408 \pm 0.0177}$ & $\bm{0.8462 \pm 0.0089}$ & $\bm{0.7986 \pm 0.0199}$ & $\bm{0.7914 \pm 0.0174}$ & $\bm{0.8160 \pm 0.0117}$ \\ \cline{1-10} 
\end{tabular}}
\label{tab:3}
\end{table*}
\renewcommand{\arraystretch}{1}

In this section, we evaluate the performances of MuViTaNet on six real-world insurance claim datasets and compare its results with state-of-the-art clinical risk prediction models to demonstrate the effectiveness of our method. Besides achieving accurate prediction, we also show the robustness of MuViTaNet in terms of interpretability.

\subsection{Datasets}
\textbf{Breast cancer cohort construction.}
We extract clinical records of female breast cancer patients from the MarketScan Commercial Claims and Encounter (CCAE) database provided by Truven Health\footnote{\url{   https://truvenhealth.com/markets/life-sciences/products/data-tools/marketscan-databases}} to construct cardiac complication risk profiling datasets. According to the previous work~\cite{liu:2019}, the records from 2012 to 2017 of de-identified patients are selected based on the following criteria.
\begin{itemize}
    \item Ages of the selected patients are from 18 to 65 at the initial diagnosis of breast cancer.
    \item The selected patients have at least six months of records and ten clinical visits before being diagnosed with breast cancer.
    \item There is no cardiac complication diagnosis until the initial diagnosis of breast cancer of the selected patients.
\end{itemize}

\textbf{Cardiac complication datasets construction.}
After construing the breast cancer cohort, we create a distinct dataset for each cardiac complication onset prediction task. In our setting, we focus on profiling the risk of developing cardiac complications in a six-month window after the initial diagnosis of breast cancer, and the positive instances are defined as patients who have cardiac complications in this window. Following previous clinical research~\cite{abdel:2019,strongman:2019}, we identify six cardiac complications including atrial fibrillation (AF), coronary artery disease (CAD), heart failure (HF), hypertension, peripheral arterial disease (PAD), and stroke. Descriptions, ICD codes, and the corresponding numbers of positive instances of these complications are shown in Table~\ref{tab:2}. The negative instances are randomly selected from the breast cancer cohort with a ratio of 3:1 compared to positive instances.

\textbf{Unlabeled dataset construction.}
The negative patients that are not selected for complication datasets are used to construct a dataset for contrastive learning. MuViTaNet leverages this dataset as additional information to improve the prediction performances of complication onset prediction tasks.

\textbf{Feature selection.}
We use the following information to profile cardiac complications for breast cancer patients.
\begin{itemize}
    \item \textit{Demographics including age and region information}. We cluster patients into three age groups (i.e., $18-44, 45-54, 55-65$) and five region groups.
    \item \textit{Clinical codes including diagnosis, procedure, and medication codes}. For diagnosis codes, all ICD-9 codes are converted to ICD-10 codes. To alleviate data sparsity, we group all diagnosis and procedure codes based on their first three characters and remove codes that appear in less than 200 patients. For medication codes, we group them by their therapeutic classes. This preprocessing step results in 1188 features.
\end{itemize}

\begin{table*}[t]
\caption{Top 10 most important clinical features (i.e., with the highest attention weights) for each cardiac complication as identified by MuViTaNet.}
\resizebox{\textwidth}{!}{
\begin{tabular}{lll}
\hline
\multicolumn{1}{c}{Atrial Fibrillation}                           & \multicolumn{1}{c}{Coronary Artery Disease}                             & \multicolumn{1}{c}{Heart Failure}                                 \\ \hline
Nonrheumatic mitral valve disorders (I34)                         & Other cardiac arrhythmias (I49)                                         & Other cardiac arrhythmias (I49)                                   \\ 
Other cardiac arrhythmias (I49)                                   & Nonrheumatic mitral valve disorders (I34)                               & Varicose veins of lower extremities (I83)                         \\ 
Complications and ill-defined heart disease (I51)                 & Varicose veins of lower extremities (I83)                               & Diseases of capillaries (I78)                                     \\ 
Paroxysmal tachycardia (I47)                                      & Diseases of capillaries (I78)                                           & Other disorders of veins (I87)                                    \\ 
Diseases of capillaries (I78)                                     & Type 2 diabetes mellitus (E11)                                          & Embolism and thrombosis (I82)                                     \\ 
Embolism and thrombosis (I82)                                     & Other peripheral vascular diseases (I73)                                & Type 2 diabetes mellitus (E11)                                    \\ 
Other conduction disorders (I45)                                  & Embolism and thrombosis (I82)                                           & Complications and ill-defined heart disease (I51)                 \\ 
Varicose veins of lower extremities (I83)                         & Hypotension (I95)                                                       & Nonrheumatic mitral valve disorders (I34)                         \\ 
Nonrheumatic aortic valve disorders (I35)                         & Other disorders of veins (I87)                                          & Other peripheral vascular diseases (I73)                          \\
Other disorders of veins (I87)                                    & Angina pectoris (I20)                                                   & Overweight and obesity (E66)                                      \\ \hline
\multicolumn{1}{c}{Hypertension}                                  & \multicolumn{1}{c}{Peripheral Arterial Disease}                         & \multicolumn{1}{c}{Stroke}                                        \\ \hline
Other cardiac arrhythmias (I49)                                   & Other cardiac arrhythmias (I49)                                         & Other cardiac arrhythmias (I49)                                   \\ 
Abnormal blood-pressure reading, without diagnosis (R03)          & Varicose veins of lower extremities (I83)                               & Nonrheumatic mitral valve disorders (I34)                         \\ 
Type 2 diabetes mellitus (E11)                                    & Diseases of capillaries (I78)                                           & Varicose veins of lower extremities (I83)                         \\ 
Nonrheumatic mitral valve disorders (I34)                         & Nonrheumatic mitral valve disorders (I34)                               & Other peripheral vascular diseases (I73)                          \\ 
Varicose veins of lower extremities (I83)                         & Other disorders of veins (I87)                                          & Embolism and thrombosis (I82)                                     \\ 
Overweight and obesity (E66)                                      & Nonspecific lymphadenitis (I88)                                         & Type 2 diabetes mellitus (E11)                                    \\ 
Diseases of capillaries (I78)                                     & Other peripheral vascular diseases (I73)                                & Other disorders of veins (I87)                                    \\ 
Other peripheral vascular diseases (I73)                          & Embolism and thrombosis (I82)                                           & Hypotension (I95)                                                 \\ 
Other disorders of veins (I87)                                    & Other noninfective disorders of lymphatic vessels (I89)                 & Pain in throat and chest (R07)                                    \\ 
Pain in throat and chest (R07)                                    & Type 2 diabetes mellitus (E11)                                          & Complications and ill-defined heart disease (I51)                 \\ \hline
\end{tabular}}
\label{tab:6}
\end{table*}

\subsection{Experimental Setup}

\textbf{Baseline Models.}
To validate the performance of the proposed model for cardiac complication risk profiling task, we compare it with several state-of-the-art models. Based on their architectures, these models are categorized into four main groups including classical model, recurrent-based model, attention-based model, and time-aware model. The details of these models are presented as follows.
\begin{itemize}
    \item \textbf{Logistic Regression (LR).} A classical model used in binary classification. To deal with insurance claim data, a patient record is converted to the count vector $\in \mathbb{Z}^{|\bm{C}|}$ whose $i^{th}$ element is the frequency of $i^{th}$ clinical code in that record, and is then fed into LR.
    \item \textbf{Random Forest (RF)~\cite{breiman:2001}.} A classical ensemble model whose prediction is the average computed from predictions of a number of decision tree classifiers. Inputs for RF are similar to LR.
    \item \textbf{Gated Recurrent Unit (GRU)~\cite{cho:2014}.} A variant of recurrent neural network (RNN) that uses gating mechanism.
    \item \textbf{Bidirectional GRU (Bi-GRU)~\cite{ljubic:2020}.} An improved version of GRU by employing an additional GRU model to learn the sequence data in reverse order.
    \item \textbf{Dipole~\cite{ma:2017}.} An attention-based model that utilizes attention mechanism over the sequence generated by Bi-GRU to learn the dependencies between visits.
    \item \textbf{RETAIN~\cite{choi:2016}.} An attention-based model that first employs a reverse RNN to process clinical records in reverse order to mimic physicians' decisions. Then two attention modules are used to identify significant visits and variables.
    \item \textbf{T-LSTM~\cite{baytas:2017}.} A time-aware model designed for handling irregularity visits in clinical records. The memory cell of LSTM is modified to capture time intervals between two consecutive visits.
    \item \textbf{Transformer~\cite{vaswani:2017}.} A fully attention-based model that uses multi-head attention mechanisms to learn the dependencies among elements in sequential data.
    \item \textbf{LSAN~\cite{ye:2020}.} An attention-based model that uses Transformer to capture global information and CNN to capture local information.
    \item \textbf{MTL Models}: We develop the MTL version for each of the aforementioned neural network-based models by employing task-specific attention and decoder over the output generated by these models.
    \item $\textbf{MuViTaNet}^{\textbf{-visit-view}}$: A variant of MuViTaNet by removing the visit-view encoder.
    \item $\textbf{MuViTaNet}^{\textbf{-feature-view}}$: A variant of MuViTaNet by removing the feature-view encoder.
    \item $\textbf{MuViTaNet}^{\textbf{-task-specific}}$: A variant of MuViTaNet by removing the task-specific attention and decoder for single-task learning (STL) setting.
    \item $\textbf{MuViTaNet}^{\textbf{-unlabeled}}$: A variant of MuViTaNet trained with labeled datasets only.
\end{itemize}

\textbf{Implementation Details.}
All neural network-based architectures are implemented by PyTorch\footnote{\url{https://pytorch.org/}}. For classical models including LR and RF, we use their Python implementations from Scikit-Learn~\cite{pedregosa:2011}. We use ADAM algorithm~\cite{kingma:2014} to optimize the prediction performances for neural network-based models. The batch size is set as $16$ for labeled datasets and $256$ for unlabeled dataset, and the initial learning rate is $0.0001$.

\textbf{Evaluation Metric.}
We conduct experiments under 5-fold cross-validation setting. $10\%$ instances from the training set are used to construct the validation set, and the results on the testing set are determined based on the best results on the validation set. The area under the receiver operating characteristic (AU-ROC) is used to measure the performances of prediction models for cardiac complication risk profiling.

\subsection{Results}

\renewcommand{\arraystretch}{1.2}

\begin{table}[t]
\centering
\caption{Average performances of MuViTaNet variants over 6 complication datasets (\textbf{F}: Feature-view, \textbf{V}: Visit-view, \textbf{L}: Labeled, \textbf{U}: Unlabeled).}
\begin{tabular}{l|cccc|c}
\hline
\multirow{2}{*}{Models} & \multicolumn{2}{c|}{Multi-view} & \multicolumn{2}{c|}{Multi-task} & \multirow{2}{*}{AU-ROC} \\ \cline{2-5}
& F & V & L & U & \\ \hline
MuViTaNet\textsuperscript{-task-specific} & \cmark & \cmark & \xmark & \xmark & $0.7385 \pm 0.0239$ \\ 
MuViTaNet\textsuperscript{-feature-view} & \xmark & \cmark & \cmark & \xmark & $0.7906 \pm 0.0286$     \\ 
MuViTaNet\textsuperscript{-visit-view}    & \cmark & \xmark & \cmark & \xmark & $0.7942 \pm 0.0248$     \\ 
MuViTaNet\textsuperscript{-unlabeled}  & \cmark & \cmark & \cmark & \xmark & $0.8102 \pm 0.0136$     \\ 
MuViTaNet & \cmark & \cmark & \cmark & \cmark & $0.8160  \pm 0.0117$ \\ \hline
\end{tabular}
\label{tab:5}
\end{table}
\renewcommand{\arraystretch}{1}

We conduct experiments to answer the following questions.
\begin{itemize}
    \item \textbf{Q1.} How accurate is MuViTaNet for cardiac complication risk profiling task comparing to previous works?
    \item \textbf{Q2.} How each component of MuViTaNet contributes to its prediction performance?
    \item \textbf{Q3.} How to effectively interpret the predictions made by MuViTaNet?
\end{itemize}

\textbf{Cardiac complication risk profiling.}
As shown in Tables~\ref{tab:3}, MuViTaNet achieves the best performances compared to other baselines for cardiac complication risk profiling task measured by AU-ROC score. Generally, it achieves an average (i.e., over six datasets) AU-ROC score of $0.8102$, which is $11\%$ better than the best previous method. Looking into each complication dataset, we also observe that MuViTaNet consistently outperforms other methods in terms of AU-ROC score. Such improvements indicate the advantage of MuViTaNet by using (1) multi-view encoder to extract comprehensive information and (2) MTL scheme to leverage information from both related labeled and unlabeled datasets to improve its prediction performance.

For baseline methods, we can observe that formulating complication risk profiling as MTL significantly improves the prediction performances of these methods. The improvements are more noteworthy for small datasets, including AF ($31\%$), CAD ($19\%$), PAD ($22\%$), and stroke ($13\%$). These results demonstrate the importance of leveraging task-related information for predicting the onset of complications. 
We also see that GRU-based models achieve slightly improved performances compared to other neural network models. For STL setting, the averaged prediction performances of deep learning models are on par with RF and are much better than LR. To investigate more, we zoom into the prediction performance for each dataset and observe that RF outperforms deep learning models for AF, CAD, PAD, and stroke datasets whose sizes are relatively small compared to HF and hypertension datasets. This result is reasonable because deep learning methods generally require large training data to achieve good prediction performance.

\textbf{Ablation study.}
To investigate the contribution of each component in MuViTaNet, we conduct an ablation study by comparing MuViTaNet with its simpler variants including MuViTaNet\textsuperscript{-visit-view}, MuViTaNet\textsuperscript{-feature-view}, MuViTaNet\textsuperscript{-task-specific}, and MuViTaNet\textsuperscript{-unlabeled} on the six aforementioned datasets. The AU-ROC scores of these models are shown in Table~\ref{tab:5}. We can observe that encoding clinical data solely by a single-view encoder is not as good as a multi-view encoder. AU-ROC score of MuViTaNet decreases to $0.7906$ (resp. $0.7942$) when only using visit-view  (resp. feature-view) encoder. This result demonstrates the necessity of aggregating information from multiple views. The performance of MuViTaNet also drops significantly when we remove the task-specific attention mechanism and decoder, which further confirms the importance of formulating complication risk profiling task as MTL with both labeled and unlabeled datasets.

\textbf{Model interpretability.}
The deployment of data-driven systems to healthcare applicants in real-world requires not only models with good prediction performance but also efficient mechanisms to interpret the automated decision to clinicians. By leveraging the multi-view multi-task architecture, our proposed model can interpret the prediction for each complication in multiple perspectives, thereby helping clinicians understand which clinical entities contribute most to the prediction.

\renewcommand{\arraystretch}{1.3}
\begin{table}[t]
\caption{Top 5 most important clinical visits and features (i.e., with the highest attention weights) for the 2 patients illustrated in Figure~\ref{fig:4}.}
\resizebox{0.48\textwidth}{!}{
\begin{tabular}{llllll}
\hline
\multicolumn{6}{c}{Positive patient from heart failure dataset}                                               \\ \hline
Visits   & Visit 9 (0.11)  & Visit 3 (0.11) & Visit 11 (0.10) & Visit 8 (0.09) & Visit 6 (0.09) \\ 
Features & 796.2 (0.26)    & 250.00 (0.25)  & 278.00 (0.12)   & 882.0 (0.05)   & 19083 (0.04)   \\ \hline
\multicolumn{6}{c}{Negative patient from hypertension dataset}                                                \\ \hline
Visits   & Visit 9 (0.11) & Visit 11 (0.11) & Visit 7 (0.10) & Visit 4 (0.10) & Visit 3 (0.09) \\ 
Features & M-174 (0.56)    & 250.00 (0.22)  & S0612 (0.13)    & J3010 (0.02)   & 82043 (0.02)   \\ \hline
\end{tabular}}
\label{tab:7}
\end{table}
\renewcommand{\arraystretch}{1}

To characterize cardiac complications, we find the most important features for each of these cardiac complications by averaging the feature-view attention weights over all positive patients for clinical features in each complication dataset. Due to the varied number of features across patients, we rescale attention weights by multiplying them with the number of features appeared in the corresponding records before averaging. Then top-10 clinical features for 6 cardiac complications are shown in Table~\ref{tab:6}. We observe that these complications share many common features such as \textbf{I34} (Nonrheumatic mitral valve disorders), \textbf{I49} (Other cardiac arrhythmias), etc. This result is reasonable because all of these complications belong to cardiovascular disease class. Moreover, many important features determined by our model are known to be clinically associated with the corresponding complications. For example, patients with type II diabetes are two to four times more likely to develop heart diseases than someone without diabetes~\cite{kenny:2019}. Obesity is another major known risk factor for heart failure and hypertension patients~\cite{mikhail:1999, ebong:2014}. Angina pectoris is the type of chest pain caused by reduced blood flow to the heart and is considered as a symptom of coronary artery disease~\cite{mosseri:1986}.

\begin{figure}[t]
\centering
\begin{subfigure}{0.52\textwidth}
    \includegraphics[width=\linewidth]{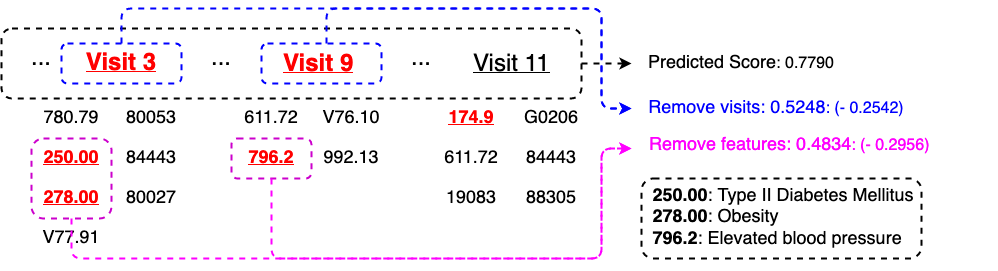}
    \caption{Positive patient from heart failure dataset}
\label{fig:4a}
\end{subfigure}\hfill
\centering
\vskip .4cm
\begin{subfigure}{0.52\textwidth}
    \includegraphics[width=\linewidth]{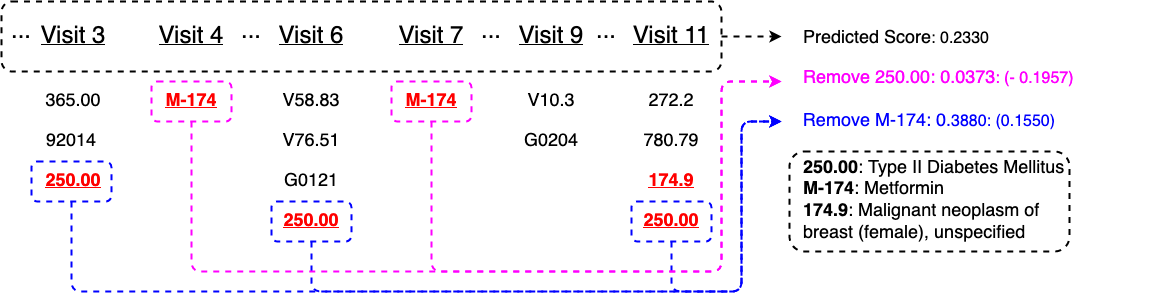}
    \caption{Negative patient from hypertension dataset}
\label{fig:4b}
\end{subfigure}\hfill
\caption{Visualization of 2 patient records (i.e., positive patient from heart failure dataset and negative patient from hypertension dataset) from breast cancer cohort. We only show important visits in clinical records due to limited space.}
\label{fig:4}
\end{figure}

\textbf{Case study.} To further investigate the interpretability of MuViTaNet, we look at two case studies to visualize the learned attention weights for finding risk factors of each complication.
The case studies include a positive patient from heart failure dataset and a negative patient from hypertension dataset. Their clinical records are illustrated in Figure~\ref{fig:4}. The most important visits and features determined by their associated attention weights from visit-view and feature-view task-specific attention components are shown in Table~\ref{tab:7}.  For the positive patient (Figure~\ref{fig:4a}), the predicted probability for heart failure onset is $0.7790$. As shown in Table~\ref{tab:7}, the visit-view attention focuses more on visits 3 and 9, which include clinical codes \textbf{250.00} (Type II diabetes mellitus) and \textbf{278.00} (Obesity) and these codes are also determined as the most important features by the feature-view attention. This result is also consistent with clinical research in which type II diabetes mellitus and obesity have been shown as the common risk factors for heart failure disease~\cite{kenny:2019, ebong:2014}, thereby demonstrating the effectiveness of MuViTaNet in capturing the correlation between risk factors and corresponding diseases. To further investigate the robustness of our model, we remove important visits and features indicating heart failure's risk factors from the patient record and predict the probability of heart failure onset based on the modified records for capturing the changes in model output. Figure~\ref{fig:4a} shows that the predicted score decreases to $0.5284$ and $0.4834$ when removing visits (3 and 9) and codes (\textbf{250.00}, \textbf{278.00}, and \textbf{796.2}) respectively. Thus, MuViTaNet is capable to focus on clinical-related visits and features when predicting onset of complications. 

Figure~\ref{fig:4b} shows a clinical record of the negative patient who has type II diabetes mellitus but is also treated by \textbf{M-174} (Metformin). Tables~\ref{tab:7} indicates that MuViTaNet pays more attention on \textbf{M-174} and \textbf{250.00} when predicting onset of hypertension. To verify whether our model can capture the relationship between disease and treatment, we remove these codes from the patient record as we did for the positive patient. Figure~\ref{fig:4b} shows that the predicted probability increases from $0.2330$ to $0.3380$ when removing Metformin (diabetes medication) and decreases to $0.0373$ when removing code \textbf{250.00} (diabetes). This result indicates that MuViTaNet considers the impact of both disease and treatment on complication development when making predictions.

\section{Conclusions}
Complication risk profiling is a crucial problem in healthcare prediction domain. In this paper, we propose a novel multi-view multi-task network (MuViTaNet) that leverages clinical data to profile multiple complications for patients. To tackle the issues of existing methods, MuViTaNet considers the record as the sequence of clinical visits and the set of clinical features, and then employs the multi-view encoder to effectively extract meaningful information from both feature-view and visit-view of the patient record. Due to the relatedness among different complications, we organize MuViTaNet as the MTL architecture in which the shared representation learned from the multi-view encoder is put into multiple task-specific attention components to learn task-specific representations for patients in both labeled and unlabeled datasets. Finally, the predicted probability for each complication onset is generated from the task-specific representation by the corresponding decoder. We evaluate the prediction performance of MuViTaNet on the insurance claim database which consists of 6 cardiac complication datasets for breast cancer survivors. The experimental results demonstrate that our proposed model outperforms other state-of-the-art models for the complication risk profiling task. More importantly, MuViTaNet provides an efficient mechanism to interpret their prediction from multiple perspectives, thereby helping clinicians to make better decisions in real-world scenarios.

\section*{Acknowledgment}
This work was funded in part by the National Science Foundation under award number CBET-2037398.

\bibliographystyle{IEEEtran}
\bibliography{ref}


\end{document}